\documentclass[review]{fcs}
\usepackage{graphicx}
\usepackage{amsmath}
\usepackage{amssymb}
\usepackage{url}

\usepackage{tabularx}
\usepackage{subfigure}
\usepackage{booktabs}
\usepackage{multirow}

\usepackage{microtype}

\title{Combat Data Shift in Few-shot Learning with Knowledge Graph}
\shorttitle{}
\author[1,2]{Yongchun Zhu}
\author[3,4]{Fuzhen Zhuang*}
\author[5]{Xiangliang Zhang}
\author[1,6]{Zhiyuan Qi}
\author[7]{Zhiping Shi}
\author[1,2]{Juan Cao}
\author[1,2]{Qing He}
\address[1]{Key Lab of Intelligent Information Processing of Chinese Academy of Sciences (CAS), Institute of Computing Technology, CAS, Beijing 100190, China}
\address[2]{University of Chinese Academy of Sciences, Beijing 100049, China}
\address[3]{Institute of Artificial Intelligence, Beihang University, Beijing 100191, China}
\address[4]{Xiamen Institute of Data Intelligence, Xiamen, Xiamen 361021, China}
\address[5]{Computer Science and Engineering, University of Notre Dame, IN 46556, USA}
\address[6]{University of California San Diego, La Jolla, CA 92093, USA}
\address[7]{College of Information Engineering \& Academy for Multidisciplinary Studies, Capital Normal University, Beijing 100089, China}

\fcssetup{
  received       = {06 20, 2021},
  accepted       = {month dd, yyyy},
  corr-email     = {*zhuangfuzhen@buaa.edu.cn},
}

\begin{abstract}
Many few-shot learning approaches have been designed under the meta-learning framework, which learns from a variety of learning tasks and generalizes to new tasks. These meta-learning approaches achieve the expected performance in the scenario where all samples are drawn from the same distributions (i.i.d. observations). However, in real-world applications, few-shot learning paradigm often suffers from data shift, i.e.,  samples in different tasks, even in the same task, could be drawn from various data distributions. Most existing few-shot learning approaches are not designed with the consideration of   data shift, and thus show  downgraded performance when data distribution shifts. However, it is   non-trivial   to address the data shift problem in few-shot learning, due to the limited number of labeled samples in each task.  
Targeting at addressing this problem, we propose a novel metric-based meta-learning framework to extract task-specific representations and task-shared representations with the help of knowledge graph. The data shift within/between tasks can thus be combated by the combination of  task-shared and  task-specific representations.
The proposed model is evaluated on popular benchmarks and two constructed new challenging datasets. The evaluation results demonstrate its remarkable performance.
\end{abstract}

\begin{document}

\section{Introduction}\label{introduction}
Few-shot learning has received increasing attention, given the urgent need for training models with limited labeled samples~\cite{vinyals2016matching,finn2017model,snell2017prototypical,sung2018learning}. The key is to address the over-fitting problem in  models that  have a large number of parameters, and get trained on a few available samples from each classification category. 
A set of few-shot learning approaches have been developed based on the meta-learning (learning-to-learn) strategy, i.e., a model is exposed to different learning tasks sampled from a training set during the training phase and is evaluated on a set of new tasks~\cite{vinyals2016matching} sampled from a test set. These meta-learning based approaches have  achieved superior performance and can be considered as a promising option to avoid overfitting~\cite{vinyals2016matching,snell2017prototypical,sung2018learning}.

It is worth noting that these approaches assume that all samples come from the same distribution (i.i.d. observations). In practice, this assumption does not often hold. For example, the popular few-shot benchmark miniImageNet~\cite{vinyals2016matching} is a subset of the large-scale dataset ImageNet~\cite{deng2009imagenet}, which actually contains   data from   diverse sources on the Internet. In other words,   samples in this dataset are drawn from different distributions, known as the shift in data distributions (a.k.a. \emph{data shift}~\cite{pan2010survey}). The problem of data shift happening between tasks (even within tasks) can downgrade the performance of few-shot learning methods designed for i.i.d. observations. Although Chen et al. realized that few-shot learning paradigm suffers from the data shift problem, they did not design a special method to solve this problem~\cite{chen2019a}.

\begin{figure*}[t!]
\centering
\begin{minipage}[b]{1\linewidth}
	\centering
	\includegraphics[width=1.0\columnwidth]{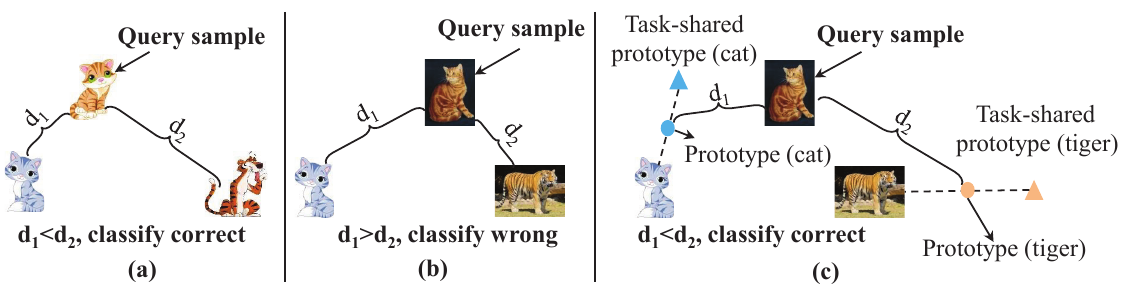}
\end{minipage}
\caption{A 2-way 1-shot task is used as an example. (a) The ideal scene of few-shot: all samples are drawn from the same distribution (cartoon). The query sample is a cartoon cat, and the support samples are a cartoon cat and a cartoon tiger. The query sample is closer to the support cartoon cat, and thus is classified correctly. (b) The real-world scene of few-shot: the samples in the same task sometimes are drawn from different distributions. The query sample is a real-world cat, and the support samples are a cartoon cat and a real-world tiger. The query sample is closer to the real-world tiger, which leads to wrong classification. (c) Our proposed solution: adjusting the task-specific prototypes by combing them with the task-shared prototype for each category,  which is learned from a given  knowledge graph. Classification decisions can be correctly made based on the new prototypes (shown as circles). 
}\label{fig:1}
\end{figure*}

In this paper, we focus on addressing the data shift problem in few-shot learning, and face the following challenges: 
\begin{itemize}
\item {\bf Intra-task data shift}. An example of intra-task data shift is shown in Figure~\ref{fig:1} (b), where samples in the same task are from different distributions. Classification error can be caused in this scenario when using models designed for i.i.d. samples, as shown in Figure~\ref{fig:1} (a) where samples in a task come from the same distribution. 
The limited number of labeled samples is the key bottleneck of few-shot learning. The quality of the support samples (labeled ones) is thus extremely important. 
When the intra-task data shift happens, the limited support samples contain insufficient information to represent its category under different distributions. 
   
\item {\bf Inter-task data shift}. The inter-task data shift occurs when different tasks are sampled from different distributions. This is mainly because the categories in different tasks could be non-overlapping, and especially the categories in new tasks are different from those used for learning. The inter-task data shift would reduce the generalization performance on new test tasks. Standard few-shot learning models also suffer from this problem. However, for solutions under the meta-learning framework, the intra-task data shift would lead the inter-task data shift more serious. 
\end{itemize}

To address these challenges, we adopt two key ideas, as shown in Figure~\ref{fig:1} (c). 
(1) Obviously, representations of each category depend on the quality of the support samples of specific tasks. In other words, the {\bf task-specific representations are unstable}. Therefore, our first idea is to incorporate auxiliary knowledge (such as word vectors) for producing a {\bf stable task-shared representation}, which would not be affected by the quality of the support samples.
(2) Due to the fact that the learning tasks and new test tasks have no overlapping categories, the second idea is to bridge all categories. A knowledge graph presenting the relationship between categories can provide semantic bridges between the categories in learning tasks and test tasks.

Inspired by a typical few-shot learning method, i.e., Prototypical Networks~\cite{snell2017prototypical}, we propose Graph Prototypical Networks (GPN), which utilizes knowledge graph via a Graph Convolutional Network (GCN) to extract task-shared representations, and uses a CNN to extract   task-specific representations from each task. 
Besides, the two representations are combined to produce a more stable class prototype, which could be less affected by the quality of samples
than standard models. Classification is then performed by simply finding the nearest class prototype of an embedded query sample. It is worth mentioning that Prototypical Networks~\cite{snell2017prototypical} is actually a special case of our approach by only using task-specific representations. In the experiments, we find that data shift has a seriously negative impact on the performance of some previous metric-based meta-learning approaches. In contrast, these approaches could have good performance in the case of no data shift. In addition, experimental results demonstrate the superiority of the proposed model.

Our contributions in this paper are summarized as follows.
\begin{itemize}
\item In this paper, we address   few-shot learning under data shift, while existing models focus on few-shot learning without the consideration of data shift. 

\item 
We propose an effective way to utilize knowledge graph to guide the learning of task-shared representations, which are combined with task-specific representations to generate appropriate class prototypes in few-shot learning.

\item To further investigate the data shift in few-shot learning, we construct two datasets, which have more explicit data shift than existing popular benchmarks based on the Office-Home~\cite{venkateswara2017deep} dataset. The evaluation results demonstrate the superior performance of the proposed GPN model. Besides, the results also indicate that current few-shot learning algorithms fail to address such data shift.

\end{itemize}

Since data shift has a seriously negative impact on the performance of some existing methods, it is important to combat data shift in few-shot learning. Note that, to the best of our knowledge, we are the first to solve data shift in few-shot learning.

\section{Related Work}
Our work is relevant to the study of 
few-shot learning, transfer learning, and graph neural network.

\textbf{Few-shot Learning:}
Early attempts on few-shot learning involved generative models with complex iterative inference strategies~\cite{fei2006one,lake2011one}. Recent trend generalized deep learning models to few-shot learning settings under the meta-learning framework~\cite{vinyals2016matching,finn2017model,snell2017prototypical}.
The work in this stream can be grouped into three clusters, i.e., metric-based, optimization-based, and parameter-generating based approaches. Metric-based meta-learning approaches learn a common feature space based on a distance metric, which enables the categories distinguished with each other~\cite{koch2015siamese,vinyals2016matching,snell2017prototypical,sung2018learning,oreshkin2018tadam,li2019lgm,allen2019infinite,liu2019learning}. Optimization-based approaches design an optimizer as the meta-learner or learns a good initialization~\cite{finn2017model,ravi2016optimization,lee2018gradient,sun2019meta}. Parameter-generating based methods usually design a parameter predicting network and use it as the meta-learner~\cite{cai2018memory,munkhdalai2017meta,munkhdalai2018rapid,peng2019few}. 
Our study falls into the  metric-based meta-learning category.

Despite effective, all the above-discussed meta-learning approaches to few-shot classification assume that the learning tasks and the new test ones are sampled from the same distribution. Chen et al.~\cite{chen2019a} found that most standard few-shot methods cannot work well under the data shift scenario. Besides, some methods~\cite{dong2018domain,guan2020few,tseng2020cross} have been proposed to solve cross-domain few-shot learning, which assumes that the training set and the test set are sampled from different domains. In practice, it is hard to distinguish boundary of domains, and the samples in the same task can be from different distributions. In this paper, we focus on addressing another challenging and practically meaningful problem: the few-shot learning under data shift, which utilizes samples from different distributions without domain labels. Note that our problem is different from cross-domain few-shot. Cross-domain few-shot has explicit data shift, while the few-shot learning under data shift aims to solve implicit data shift. 

Note that KTN~\cite{peng2019few} is the most related work which also utilizes a knowledge graph in few-shot learning. Hence, we compare our method with KTN in detail as following: (1) The training frameworks of the two methods are entirely different. With the meta-learning framework, in the training stage, each iteration of our method uses a few-shot task for training, e.g., in 5-way 1-shot task, a training task contains 5 support samples and several query samples from 5 classes. On the contrary, KTN does not use the meta-learning framework, and it randomly samples a batch from the training set for one iteration as standard supervised learning does. (2) KTN is in the family of parameter-generating based methods while our method is in the family of metric-learning solutions (The difference between the two families can be found above). (3) Our method exploits KG and GCN for learning more stable representations of categories, while KTN uses KG and GCN for predicting the parameters of classifiers, which is similar to some zero-shot methods~\cite{wang2018zero-shot,kampffmeyer2018rethinking}. (4) We focus on combating data shift in few-shot learning while KTN did not consider data shift. In addition, the experimental results reveal that our method outperforms KTN for few-shot learning under data shift.

\textbf{Transfer Learning:} Learning a discriminative model in the presence of the shift between training and test distributions is known as transfer learning~\cite{pan2010survey,zhuang2020comprehensive,wang2021generalizing}. Recently, deep transfer approaches have made significant achievements in improving the performance of learners~\cite{zhuang2015supervised,ganin2016domain,wang2017balanced,wang2018visual,zhu2019multi,xi2020domain,zhu2021transfer}. These approaches can be divided into several classes including adversarial discriminative approaches~\cite{ganin2016domain,tzeng2015simultaneous,tzeng2017adversarial}, discrepancy reducing approaches~\cite{long2015learning,long2016deep,zhu2020deep}, adversarial generative approaches~\cite{bousmalis2017unsupervised,hoffman2017cycada}, and reconstruction approaches~\cite{ghifary2016deep}. These approaches work under the setting that the training set and the test set have shared categories. However, in few-shot learning, the categories of the training set (learning tasks) and the test set (new test tasks) are non-overlapping. 
Therefore, these approaches are not feasible for few-shot learning under data shift. 

\textbf{Graph Neural Network:}
Graph neural network (GNN) is a new type of neural network for learning over graphs~\cite{scarselli2009graph,bruna2013spectral,henaff2015deep}. 
Garcia et al.~\cite{garcia2017few} used GNN on the top of CNN to capture the invariances of the tasks for few-shot learning. In contrast, Kim et al.~\cite{kim2019edge} proposed EGNN that learns to predict the edge-labels rather than the node-labels on the graph. Both of these two methods~\cite{garcia2017few,kim2019edge} construct graphs with the features extracted from images. In our proposed approach, we use an existing knowledge graph. Graph convolutional networks (GCN) are a class of GNN, which are based on local graph operators~\cite{bruna2013spectral,defferrard2016convolutional,kipf2016semi}. Some approaches~\cite{wang2018zero-shot,kampffmeyer2018rethinking,zhuo2019unsupervised,peng2019few} use GCN to predict the parameters of classifiers. We use the same GCN architecture as~\cite{kampffmeyer2018rethinking,zhuo2019unsupervised}. Note that we also use GCN for another purpose, i.e., learning task-shared representations of each category.

\textbf{Knowledge Graph:} Knowledge Graph (KG) has been applied in computer vision tasks, e.g., object recognition~\cite{salakhutdinov2011learning}, visual question answering~\cite{wu2016ask}, zero-shot learning~\cite{wang2018zero-shot,kampffmeyer2018rethinking}. The most related work is the few-shot learning with KG~\cite{peng2019few}, and the differences between our method and KTN have been listed above.

\section{Model}
In this section, we first introduce the problem setting, and then propose a new model named GPN to learn task-specific and task-shared representations, respectively, which are combined to produce more stable representations.
\subsection{Problem Setting}
Formally, we have a training set $D_{\text{train}}$ and a test set $D_{\text{test}}$, and their class labels do not overlap.
Existing standard few-shot learning approaches assume that all samples in the training set $D_{\text{train}}$ and the test set $D_{\text{test}}$ are drawn from the same distribution. In contrast, we assume that the samples could be drawn from different distributions, and we do not know which distribution a sample comes from.
The goal is to build a classifier for $D_{\text{test}}$ by training a model on $D_{\text{train}}$ in the scenario of few-shot learning under data shift.

To achieve this goal, a model is trained on a set of $N$-way $K$-shot learning tasks (i.e., going through a meta-learning phase). An $N$-way $K$-shot learning task is constructed by randomly selecting $N$ classes from the ones of the training set $D_{\text{train}}$ and then generating two sets including: 1) a support set $S = \{ (\mathbf{x}_1, y_1), \cdots, (\mathbf{x}_{N \times K}, y_{N \times K}) \}$ containing $K$ samples in each of the $N$ selected classes; and 2) a query set $Q$ containing a number of distinct samples whose classes are the $N$ selected classes. In the training process, for a constructed learning task, the support set $S$ is fed to the model, and then the parameters of the model are updated to minimize the loss of its predictions for samples in the query set $Q$. In the test process, we sample $N$-way $K$-shot test tasks along with a support set $S$ and a query set $Q$ from the test set $D_{\text{test}}$, which is similar to the constructions of the training tasks. The labels of   samples in the query set are predicted with the help of the support set.

\begin{figure*}[t!]
	\centering
	\begin{minipage}[b]{1\linewidth}
		\centering
		\includegraphics[width=1.0\linewidth]{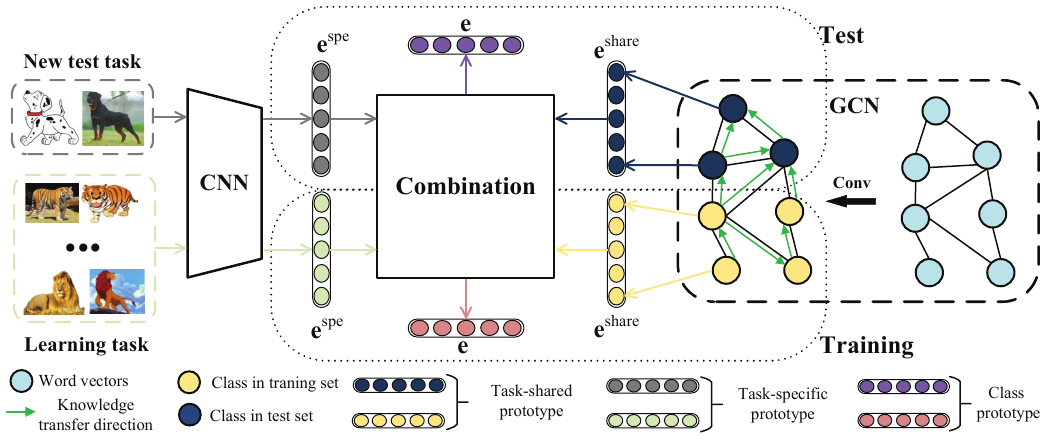}
	\end{minipage}
	\caption{The proposed GPN framework for few-shot classification under data shift, which consists of a CNN (with images as input) and a GCN (with word vectors as input) for extracting task-specific and task-shared prototypes, respectively. The class prototype used for classification is the combination of the two kinds of prototypes. Note that, in a learning task, the data consist of images from different distributions. The green arrow shows that the GCN could transfer the knowledge contained in the classes of the training set to the classes of the test set.}\label{fig:network}
\end{figure*}

\subsection{Task-specific Representations}
Inspired by prototypical networks~\cite{snell2017prototypical}, we use the support set to compute a $V$-dimensional representations $\mathbf{e} \in \mathbb{R}^V$ for each class (a.k.a., \emph{prototype}) through an embedding function $f(\cdot)$. For each task, a prototype $\mathbf{e}_n$ is defined as the mean vector of the embeddings of the support samples from the same class $n$, i.e.,
\begin{equation}
    \mathbf{e}_n = \frac{1}{|S_n|}\sum_{(\mathbf{x}_i, y_i)\in S_n} f(\mathbf{x}_i),\label{task-specific}
\end{equation}
where $S_n \in S$ is the subset (with the label of class $n$) of the support set $S$. According to Equation (\ref{task-specific}), we could find the quality of $\mathbf{e}_n$ depends heavily on the support set $S$ of specific tasks.
In other words, the quality of prototype $\mathbf{e}$ of the same category varies from one task to another, which contains task-specific information. For convenience, we use $\mathbf{e}_n^{\text{spe}}$ to denote the task-specific prototype extracted by $f(\cdot)$ for class $n$.


\subsection{Task-shared Representations}
To solve the intra-task and inter-task data shift, the two key ideas are learning stable task-shared representations and bridging categories as mentioned above.
Exploiting the knowledge graph allows us to capture (1) semantic information of categories which is stable, and (2) semantic relationships that bridge all categories. 
Even if samples are from different distributions, they often share the auxiliary knowledge encoded in word vectors of categories as well as the prior semantic relationships between classes. Therefore, the knowledge graph can drive the learning of the stable task-shared representations and bridge the gaps between categories in learning tasks and test tasks. In order to integrate word vectors of categories and knowledge graph structure, we adopt GCN~\cite{kipf2016semi}, which is a state-of-the-art attributed graph embedding model.

A graph (with $L$ nodes) is constructed at first, and each node denotes a $J$-dimensional vector representing a distinct concept (the name of a class). Our goal is to transfer the knowledge contained in categories with sufficient labeled samples (in $D_{\text{train}}$) to learn the categories with few labeled samples (in $D_{\text{test}}$). Auxiliary nodes from WordNet are exploited to construct paths from categories in $D_{\text{train}}$ to the categories in $D_{\text{test}}$. Each node is initialized with the word vector of the class label name. The relationships between the classes in the knowledge graph (e.g., WordNet) are encoded in the form of symmetric adjacency matrix $A \in \mathbb{R}^{L \times L}$ including self-loops. A propagation rule is employed to perform convolutions on the graph, i.e.,
\begin{equation}
    H^{(l+1)} = \sigma (D^{-1}AH^{(l)}\Theta^{(l)}),
\end{equation}
where $H^{(l)}$ represents the activations in the $l$-th layer, $\Theta^{(l)}$ denotes the trainable weight matrix of layer $l$, $D \in \mathbb{R}^{L \times L}$ is a degree matrix with $D_{ii} = \sum_j A_{ij}$. For the first layer, $H^{(0)} \in \mathbb{R}^{L \times J}$ is composed of $L$ word vectors. The GCN $g(\cdot)$ is trained to predict the task-shared prototype of all categories, i.e., $\mathbf{e}^{\text{share}} = g(H^{(0)})$, where $\mathbf{e}^{\text{share}}$ denotes the task-shared prototype and is made of $\mathbf{e}^{\text{share}}_n$ (with $n$ denoting the class).

\subsection{Graph Prototypical Networks}
In our proposed Graph Prototypical Networks (GPN), which is shown in Figure~\ref{fig:network}, CNN $f(\cdot)$ is used to extract task-specific prototypes $\mathbf{e}^{\text{spe}}$ that would be affected by the quality of the support set $S$. And the word vector corresponding to each knowledge graph node is used as the input of GCN $g(\cdot)$ to extract the task-shared prototypes $\mathbf{e}^{\text{share}}$. The task-specific and task-shared prototypes are combined to produce   stable prototypes of each class, i.e.,
\begin{equation}
    \mathbf{e}_n = \lambda \cdot \mathbf{e}^{\text{spe}}_n + ( 1 - \lambda ) \cdot \mathbf{e}^{\text{share}}_n,
\end{equation}
where $\lambda \in [0, 1]$ is a trade-off parameter between the task-specific and the task-shared representations. For a query example $\mathbf{x}$, GPN predicts its class label based on the softmax operations over the distances between $f(\mathbf{x})$ and the prototypes in the embedding space, i.e.,
\begin{equation}
    P(y = n | \mathbf{x}) = \frac{\exp( -d( f(\mathbf{ x }), \mathbf{e}_n ) ) }{ \sum_{ n'  } \exp(  -d( f(\mathbf{ x }), \mathbf{e}_{ n'  } )  )  },
\end{equation}
where $d(\cdot, \cdot)$ represents the Euclidean distance. The overall loss function is defined as the cross-entropy loss on the query set $Q$, i.e., 
\begin{equation}
    J = -\frac{1}{|Q|} \sum_{ (\mathbf{x}_i, y_i) \in Q } \log P(y = y_i | \mathbf{x}_i ).\label{crossentropy}
\end{equation}
The training process consists of several steps, i.e., minimizing the average loss, iterating over learning tasks, and performing a gradient descent update for each task. The generalization performance is measured on new test tasks, where the samples are from the classes in $D_{\text{test}}$ rather than $D_{\text{train}}$. For each test task, GPN is also used to calculate all $\mathbf{e}_n$ from the provided support set $S$ and to predict the class of each query example $\mathbf{x}$ as the one with the highest probability, i.e., $y=\arg \max_n P(y = n | \mathbf{x})$.

\section{Experiments}
In this section, we conduct extensive experiments to demonstrate the impact of data shift on few-shot learning and the superiority of the proposed model.

\subsection{Datasets}

\textbf{miniImageNet} is originally proposed by~\cite{vinyals2016matching}, which consists of 60,000 colour images in total from 100 classes (i.e., 600 samples per class). We follow the split method in~\cite{ravi2016optimization}, i.e., using 64, 16, and 20 classes for training, validation, and testing, respectively. The 16 validation classes are used to monitor the generalization performance. In \cite{chen2019a}, Chen et al. have discovered the data shift problem in miniImageNet.

\textbf{Office-Home}~\cite{venkateswara2017deep} is a dataset consisting of 15,588 images, which has been widely adopted in cross-domain classification~\cite{long2018conditional,zhu2019aligning}. The images come from four different domains, which include Artistic, Clip-Art, Product, and Real-World images. For each domain, the dataset contains 65 object categories with the images collected from office and home settings. There is some great discrepancy between these four domains, and thus they can be used to construct and simulate various kinds of data shift scenarios. Some images in this dataset are shown in Figure~\ref{fig:2a}.

\textbf{Easy-Office-Home} and \textbf{Hard-Office-Home} are our constructed two datasets for few-shot classification under data shift. These two datasets are the subsets of Office-Home. As mentioned before, the four domains in Office-Home have the same 65 categories. Therefore, the two datasets are constructed as follows. First, the 65 classes are randomly split into three parts, i.e., 38 for training, 12 for validation, and the rest 15 for test. Then, one of the four domains is chosen as the target domain with the other three domains as the auxiliary domains. The validation set and test set of all settings use the images of the 12 validation classes and the 15 test classes from the target domain, respectively. To further investigate the influence of data shift, we introduce three settings: 
(1) \textbf{tgt}: only using samples belonging to  the 12 training classes in {target} domain for training; (2) \textbf{src+tgt}: using all the auxiliary domain samples (from 38 training classes) and the target domain samples (from 12 training classes); (3) \textbf{fulltgt}: using all the samples of 38 training classes in the target domain. Note that all samples under the settings of tgt and fulltgt come from the same domain (distribution), which could be regarded as no existing data shift. Besides, under {\bf src+tgt} setting, samples of the learning tasks could come from four domains, causing {\bf  intra-task data shift}. Meanwhile, samples of the test tasks only come from a single target domain, resulting in {\bf  inter-task data shift}.

According to the results of the cross-domain classification method in~\cite{long2018conditional}, we find that the average accuracy of cross-domain tasks with Product and Clip Art as the target domains are $74.1\%$ and $52.8\%$, respectively. This result indicates that the cross-domain task on the Product domain is easier than the task on the Clip Art domain. Therefore, we respectively use Product and Clip Art domains as the target domains in Easy-Office-Home and Hard-Office-Home datasets with the rest three domains as the auxiliary domains.

\begin{table*}[!t]
  \centering
  \small
  \caption{\centering  The 5-way few-shot classification accuracy (\%) of the models on miniImageNet, where the models are divided into three groups, i.e., optimization-based (\textbf{O}), parameter-generating based (\textbf{P}), and metric-based (\textbf{M}) ones.}
    \begin{tabular}{p{1cm}<{\centering}p{2.5cm}<{\centering}p{2.5cm}<{\centering}p{2.5cm}<{\centering}p{2.5cm}<{\centering}}
    \toprule
          & Model & Embedding & 1-shot & 5-shot \\
    \midrule
    \multirow{4}[2]{*}{\textbf{O}} & MAML~\cite{finn2017model}  & ConvNet & 48.7$\pm$0.8 & 55.3$\pm$0.7 \\
          & MetaGAN~\cite{zhang2018metagan} & ResNet-12 & 52.7$\pm$0.6 & 68.6$\pm$0.7 \\
          & Bilevel~\cite{franceschi2018bilevel} & ResNet-12 & 50.5$\pm$0.9 & 64.5$\pm$0.7 \\
          & MTL~\cite{sun2019meta}   & ResNet-12 & \underline{61.2}$\pm$1.8 & 75.5$\pm$0.8 \\
    \midrule
    \multirow{3}[2]{*}{\textbf{P}} & MetaNet~\cite{munkhdalai2017meta} & ConvNet & 49.2$\pm$1.0 & - \\
          & MM-Net~\cite{cai2018memory} & ConvNet & 53.4$\pm$0.5 & 66.9$\pm$0.4 \\
          & adaNet~\cite{munkhdalai2018rapid} & ResNet-12 & 56.9$\pm$0.6 & 71.9$\pm$0.6 \\
          & KTN~\cite{peng2019few} & ResNet-12 & 61.4$\pm$0.7 & 70.2$\pm$0.6 \\
    \midrule
    \multirow{9}[2]{*}{\textbf{M}} & MatchNet~\cite{vinyals2016matching} & ConvNet & 43.4$\pm$0.8 & 55.3$\pm$0.7 \\
          & ProtoNet~\cite{snell2017prototypical} & ConvNet & 49.4$\pm$0.8 & 68.2$\pm$0.7 \\
          & RelationNet~\cite{sung2018learning} & ConvNet & 50.4$\pm$0.8 & 65.3$\pm$0.7 \\
          & GNN~\cite{garcia2017few} & ConvNet & 50.3$\pm$0.4 & 66.4$\pm$0.6 \\
          & IMP~\cite{allen2019infinite} & ConvNet & 49.6$\pm$0.8 & 68.1$\pm$0.8 \\
          & TADAM~\cite{oreshkin2018tadam} & ResNet-12 & 58.5$\pm$0.3 & \textbf{76.7}$\pm$0.3 \\
          & CAML~\cite{jiang2018learning} & ResNet-12 & 59.2$\pm$1.0 & 72.4$\pm$0.7 \\
          & \textbf{GPN~(Ours)} & ConvNet &  56.1$\pm$0.7 & 68.4$\pm$1.4  \\
          & \textbf{GPN~(Ours)} & ResNet-12 &  \textbf{63.0}$\pm$0.8     &  \underline{75.6}$\pm$0.4 \\
    \bottomrule
    \end{tabular}%
  \label{tab:miniimagenet}%
\end{table*}%



\subsection{Setup}

\textbf{Baselines:} On miniImageNet, we compare the proposed model with the state-of-the-art standard few-shot learning methods: (1) Optimization-based (\textbf{O}): MAML~\cite{finn2017model}, MetaGAN~\cite{zhang2018metagan}, Bilevel~\cite{franceschi2018bilevel}, MTL~\cite{sun2019meta}; (2) Parameter-generating (\textbf{P}): MetaNet~\cite{munkhdalai2017meta}, MM-Net~\cite{cai2018memory}, adaNet~\cite{munkhdalai2018rapid}, KTN~\cite{peng2019few}; (3) Metric-based: MatchNet~\cite{vinyals2016matching}, ProtoNet~\cite{snell2017prototypical}, RelationNet~\cite{sung2018learning}, GNN~\cite{garcia2017few}, IMP~\cite{allen2019infinite}, TADAM~\cite{oreshkin2018tadam}, CAML~\cite{jiang2018learning}. For all the baselines, we cite the results from their papers. 

On datasets of Easy-Office-Home and Hard-Office-Home, which suffer from obvious data shift, we compare our model against the most related work KTN~\cite{peng2019few} and several typical metric-based meta-learning baselines, which include MatchNet~\cite{vinyals2016matching}, ProtoNet~\cite{snell2017prototypical} and RelationNet~\cite{sung2018learning}, for few-shot classification in a more comprehensive manner. 
Due to the lack of prior models addressing the data shift problem in few-shot learning, some transfer learning methods are incorporated into ProtoNet~\cite{snell2017prototypical} to generate competitive baselines for comparison. Specifically, three typical transfer models are adopted, i.e., DAN~\cite{long2015learning}, DANN~\cite{ganin2016domain}, D-CORAL~\cite{sun2016deep}. And the corresponding baselines are named ProtoNet + DAN, ProtoNet + DANN, and ProtoNet + D-CORAL. These baselines are trained with two  repeated steps:   (1) training a learning task in the same way as ProtoNet does; (2) choosing two batches of samples in training set and aligning their features as transfer methods do (aligning the distributions for different tasks). 

\begin{table*}[!th]
	\centering
	\caption{\centering The few-shot classification accuracy on Easy-Office-Home.} \label{tab:easyOfficeHome}
	\begin{tabular}{@{}ccccccc@{}}
		\toprule
		& &  \multicolumn{2}{c}{$\mathbf{5}$\textbf{-Way}} & \multicolumn{2}{c}{$\mathbf{10}$\textbf{-Way}} & \\
		\textbf{Setting} & \textbf{Model} & $\mathbf{1}$\textbf{-shot} & $\mathbf{5}$\textbf{-shot} & $\mathbf{1}$\textbf{-shot} & $\mathbf{5}$\textbf{-shot} &
		\textbf{avg} \\
		\midrule
		& MatchNet & 36.4\%$\pm$2.0 & 38.8\%$\pm$1.7 & 19.5\%$\pm$2.1 & 22.1\%$\pm$0.4 & 29.2\% \\
		\textbf{tgt}& ProtoNet & 41.9\%$\pm$1.3 & 59.6\%$\pm$0.4 & 29.2\%$\pm$2.1 & 49.5\%$\pm$0.8 & 45.1\% \\
		&RelationNet & 35.5\%$\pm$1.6 & 50.3\%$\pm$2.4 & 23.3\%$\pm$1.4 & 33.1\%$\pm$3.6 & 35.6\% \\
		\midrule
		& MatchNet & 33.1\%$\pm$0.6 & 35.7\%$\pm$2.9 & 19.0\%$\pm$0.4 & 19.3\%$\pm$1.5 & 26.8\% \\
		& ProtoNet & 53.4\%$\pm$1.4 & 71.2\%$\pm$0.3 & \underline{41.1}\%$\pm$0.4 & 60.2\%$\pm$1.4 & \underline{56.4}\% \\
		&RelationNet & 37.2\%$\pm$0.6 & 57.3\%$\pm$1.9 & 30.0\%$\pm$2.5 & 45.9\%$\pm$1.3 & 42.6\% \\
		\textbf{src+tgt}& ProtoNet + DAN & 52.1\%$\pm$0.6 & 69.6\%$\pm$1.8 & 39.5\%$\pm$0.6 & 61.2\%$\pm$1.9 & 55.6\% \\
		&ProtoNet + DANN & 52.7\%$\pm$0.9 & 71.0\%$\pm$0.6 & 38.9\%$\pm$1.1 & \underline{61.4}\%$\pm$2.1 & 56.0\% \\
		&ProtoNet + D-CORAL & 50.5\%$\pm$0.9 & 69.9\%$\pm$1.2 & 39.5\%$\pm$0.7 & 59.3\%$\pm$0.3 & 54.8\% \\
		& KTN & \underline{54.3}\%$\pm$0.8 & \underline{72.6}\%$\pm$2.2 & 38.5\%$\pm$1.4 & 58.8\%$\pm$0.6 & 56.0\% \\
		&\textbf{GPN} & \textbf{57.4}\%$\pm$0.9 & \textbf{73.1}\%$\pm$0.3 & \textbf{45.6}\%$\pm$0.9 & \textbf{62.2}\%$\pm$0.6 & \textbf{59.6}\%\\
		\midrule
		& MatchNet & 36.5\%$\pm$2.8 & 43.3\%$\pm$2.6 & 23.5\%$\pm$1.0 & 26.8\%$\pm$0.9 & 32.5\% \\
		\textbf{fulltgt}& ProtoNet & 54.6\%$\pm$1.6 & 72.4\%$\pm$1.8 & 41.3\%$\pm$1.0 & 63.8\%$\pm$2.0 & 58.0\% \\
		& RelationNet & 42.1\%$\pm$0.7 & 59.0\%$\pm$1.2 & 31.5\%$\pm$2.1 & 47.8\%$\pm$1.3 &  45.1\% \\
		\bottomrule
	\end{tabular}
\end{table*}

\textbf{Implementation Details:} for models implemented on miniImageNet, following~\cite{snell2017prototypical,oreshkin2018tadam}, our GPN use ConvNet and ResNet-12 as the embedding modules, respectively. The size of the input images is 84 $\times$ 84. SGD Optimizer~\cite{sutskever2013importance} is adopted for optimization. The initial learning rate is set to 0.1 and the momentum coefficient is fixed to 0.9. Each model is trained by 40,000 iterations and the learning rate is annealed by a factor of ten every 10,000 iterations. 

\begin{table*}[t]
	\centering
	\caption{\centering Few-shot classification accuracy on Hard-Office-Home.} \label{tab:hardOfficeHome}
	\begin{tabular}{@{}ccccccc@{}}
		\toprule
		& &  \multicolumn{2}{c}{$\mathbf{5}$\textbf{-Way}} & \multicolumn{2}{c}{$\mathbf{10}$\textbf{-Way}} & \\
		\textbf{Setting} &\textbf{Model} & $\mathbf{1}$\textbf{-shot} & $\mathbf{5}$\textbf{-shot} & $\mathbf{1}$\textbf{-shot} & $\mathbf{5}$\textbf{-shot} &
		\textbf{avg} \\
		\midrule
		& MatchNet & 27.1\%$\pm$1.7 & 30.1\%$\pm$2.3 & 16.4\%$\pm$1.5 & 20.9\%$\pm$1.5 & 23.6\% \\
		\textbf{tgt}& ProtoNet & 38.5\%$\pm$0.3 & 51.4\%$\pm$1.1 & 25.0\%$\pm$0.3 & 40.9\%$\pm$1.1 & 39.0\% \\
		&RelationNet & 31.8\%$\pm$0.6 & 43.2\%$\pm$0.7 & 21.1\%$\pm$0.5 & 27.8\%$\pm$1.6 & 31.0\% \\
		\midrule
		&MatchNet & 24.9\%$\pm$1.0 & 26.9\%$\pm$0.8 & 14.6\%$\pm$0.4 & 15.8\%$\pm$1.3 & 20.6\% \\
		&ProtoNet & 44.9\%$\pm$0.5 & \textbf{60.7}\%$\pm$2.5 & 31.2\%$\pm$0.9 & \underline{49.2}\%$\pm$1.1 & \underline{46.5}\% \\
		&RelationNet & 28.9\%$\pm$0.2 & 46.6\%$\pm$1.4 & 20.2\%$\pm$1.5 & 37.7\%$\pm$2.6 & 33.4\% \\
		\textbf{src+tgt}&ProtoNet + DAN & 43.4\%$\pm$1.2 & 59.2\%$\pm$0.5 & 30.2\%$\pm$1.1 & 46.7\%$\pm$3.1 & 44.9\% \\
		&ProtoNet + DANN & 44.0\%$\pm$0.7 & 59.0\%$\pm$1.8 & 31.0\%$\pm$1.2 & 46.9\%$\pm$1.1 & 45.2\% \\
		&ProtoNet + D-CORAL & 43.9\%$\pm$1.2 & 59.2\%$\pm$1.0 & 30.9\%$\pm$0.5 & 46.2\%$\pm$1.9 & 45.1\% \\
		& KTN & \underline{47.3}\%$\pm$0.4 & 59.9\%$\pm$0.7 & \underline{32.4}\%$\pm$0.8 & 45.4\%$\pm$0.5 & 46.3\% \\
		&\textbf{GPN} & \textbf{49.7}\%$\pm$0.8 & \underline{60.5}\%$\pm$0.8 & \textbf{36.5}\%$\pm$0.8 & \textbf{50.1}\%$\pm$0.9 & \textbf{49.2}\%\\
		\midrule
		& MatchNet & 26.7\%$\pm$2.2 & 39.3\%$\pm$3.7 & 17.3\%$\pm$0.5 & 25.0\%$\pm$2.2 & 27.1\% \\
		\textbf{fulltgt}&ProtoNet & 46.9\%$\pm$1.3 & 64.4\%$\pm$1.2 & 33.2\%$\pm$0.8 & 51.8\%$\pm$1.7 & 49.1\% \\
		&RelationNet & 38.2\%$\pm$1.9 & 51.9\%$\pm$2.0 & 27.4\%$\pm$0.3 & 41.0\%$\pm$1.1 & 39.6\% \\
		\bottomrule
	\end{tabular}
\end{table*}

In terms of Easy-Office-Home and Hard-Office-Home, for all models, the size of the input images is 224 $\times$ 224. In the training process, horizontal flip and random crop operations are performed for data augmentation. Previous standard few-shot models are usually trained from scratch, which costs a lot of time. Given that the two datasets may not contain sufficient training data to produce a satisfying model, we employ the ResNet-18~\cite{he2016deep} and fine-tune all the convolutional and pooling layers, which are pre-trained on ImageNet, via back-propagation. In order to conduct a fair comparison, all the baselines and the proposed model use the same architecture, i.e., ResNet-18, as the embedding function. All the models are trained by 5,000 iterations with a learning rate of 0.001 and 0.005 for embedding function and extra function (the relation module in RelationNet~\cite{sung2018learning} and the adopted GCN) using Adam~\cite{kingma2014adam}, respectively. In addition, the learning rate is decreased by half every 2000 iterations. 

For all the datasets, the test models are chosen based on the classification accuracy on the validation set. 
The classification accuracy is computed for all the models by averaging over 1,000 randomly generated new tasks from the test set. All the $N$-way (with $N=5$~and~$10$) $K$-shot tasks (with $K = 1$~and~$5$) are conducted with $K$ support samples and 15 query samples in each category. All the models are implemented in Pytorch~\cite{paszke2017automatic} and the average classification accuracy is reported with 95\% confidence intervals of three random trials.

In the meta-learning framework for few-shot learning, we have no idea about which category will appear in the meta-test stage. Thus, we cannot utilize the (unknown) prior information about categories in the meta-test stage to adjust KG. In both KTN and the proposed GPN, for covering all categories in both the meta-training and meta-test stages, a large KG is constructed based on WordNet~\cite{miller1995wordnet} that contains sufficient categories. MiniImageNet and OfficeHome are two subsets of the ImageNet 2012 1K dataset. All categories in ImageNet 2012 1K are used to construct the KG. These categories are connected based on the relationship of hypernym in WordNet. The nodes include all the categories as well as their children and ancestors. The construction of the KG follows~\cite{kampffmeyer2018rethinking,wang2018zero-shot}, and the constructed KG contains about 30K nodes covering sufficient categories. 
The word vectors, which correspond to the categories, are extracted via the GloVe text model~\cite{pennington2014glove} that is trained on Wikipedia. The dimension of the word vectors, which are the input of GCN, is set to 300. The GCN model consists of two layers with the feature dimension of 2048, 
and the dimension of its final output is determined by the dimension of outputs in the last layer of the CNN architecture (ConvNet, ResNet-12, ResNet-18). In 1-shot tasks, $\lambda$ is set to 0.5. And in 5-shot tasks, since more labeled samples would provide more task-specific information, $\lambda$ is set to 0.7.

\subsection{Results}
The results of miniImageNet, Easy-Office-Home, and Hard-Office-Home are respectively shown in Table~\ref{tab:miniimagenet}, ~\ref{tab:easyOfficeHome} and ~\ref{tab:hardOfficeHome}, where the best results are marked in bold, and the second best results are underlined. The experimental results reveal some insightful points from two aspects: the effectiveness of the proposed GPN and the influence of data shift. Note that, in Table~\ref{tab:miniimagenet}, the results of all baselines are extracted from their original paper. All results in Table~\ref{tab:easyOfficeHome} and ~\ref{tab:hardOfficeHome} are obtained by us.

\textbf{The Effectiveness of GPN:} 
\begin{itemize}
    \item[(1)] The proposed GPN could achieve remarkable results on the popular benchmark miniImageNet.
    \item[(2)] Under src+tgt setting, GPN almost outperforms all the compared models on all the tasks. In particular, GPN substantially improves the average accuracy by a large margin (about $3$\%) on the constructed two datasets over the best baseline. The existing results demonstrate the effectiveness of the proposed GPN for handling such few-shot under the data shift problem.
    \item[(3)] Due to the less shot condition, traditional few-shot learning approaches are more easily affected by the quality of samples. Under this circumstance, utilizing the additional task-shared information could greatly help the task-specific representations to achieve stability and reliability, which is evidenced by the phenomenon that the proposed GPN significantly outperforms the baselines in terms of the classification performance.
    \item[(4)] Both KTN~\cite{peng2019few} and our GPN utilize knowledge graph for few-shot learning, and our GPN outperforms KTN on most tasks, which verifies that our GPN is more effective than KTN. The main reason would be that KTN exploits KG and GCN as most zero-shot methods~\cite{wang2018zero-shot,kampffmeyer2018rethinking} do, which could be unsuitable for few-shot learning. On the contrary, our GPN utilizes the KG and GCN based on the few-shot framework, which is more suitable for few-shot learning. 
\end{itemize}

\begin{table*}[t!]
	\centering
	\caption{\centering Ablation Study: Few-shot classification accuracy on Easy-Office-Home.} \label{ablation}
	\begin{tabular}{@{}cccccc@{}}
		\toprule
		&  \multicolumn{2}{c}{$\mathbf{5}$\textbf{-Way}} & \multicolumn{2}{c}{$\mathbf{10}$\textbf{-Way}} & \\
		\textbf{Model} & $\mathbf{1}$\textbf{-shot} & $\mathbf{5}$\textbf{-shot} & $\mathbf{1}$\textbf{-shot} & $\mathbf{5}$\textbf{-shot} &
		\textbf{avg} \\
		\midrule
		\textbf{GPN (rand)} & 43.3\%$\pm$4.9 & 68.0\%$\pm$3.6 & 39.4\%$\pm$4.1 & 59.8\%$\pm$2.1 & 52.6\%\\
		\textbf{GPN (fc)} & 54.4\%$\pm$2.6 & 71.3\%$\pm$0.7 & 43.8\%$\pm$1.2 & 61.6\%$\pm$2.2 & 57.8\%\\
		\textbf{GPN} & \textbf{57.4}\%$\pm$0.9 & \textbf{73.1}\%$\pm$0.3 & \textbf{45.6}\%$\pm$0.9 & \textbf{62.2}\%$\pm$0.6 & \textbf{59.6}\%\\
		\bottomrule
	\end{tabular}
\end{table*}

\begin{figure*}[t!]
	\centering
	\subfigure[Images of Different Domains]{
		\begin{minipage}[b]{0.40\linewidth}
			\centering
			\includegraphics[width=.95\columnwidth]{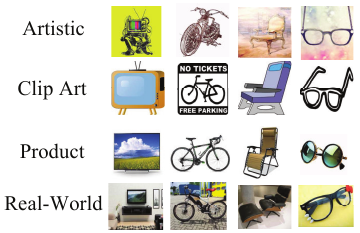}
			\label{fig:2a}
		\end{minipage}
	}
	\subfigure[Feature Visualization]{
		\begin{minipage}[b]{0.26\linewidth}
			\centering
			\includegraphics[width=.95\columnwidth,height=.95\columnwidth]{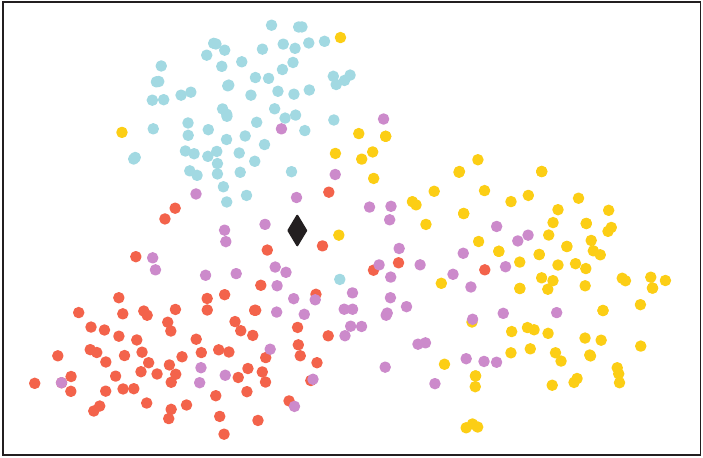}
			\label{fig:2b}
		\end{minipage}
	}
	\subfigure[Accuracy w.r.t $\lambda$]{
		\begin{minipage}[b]{0.26\linewidth}
			\centering
			\includegraphics[width=.95\columnwidth]{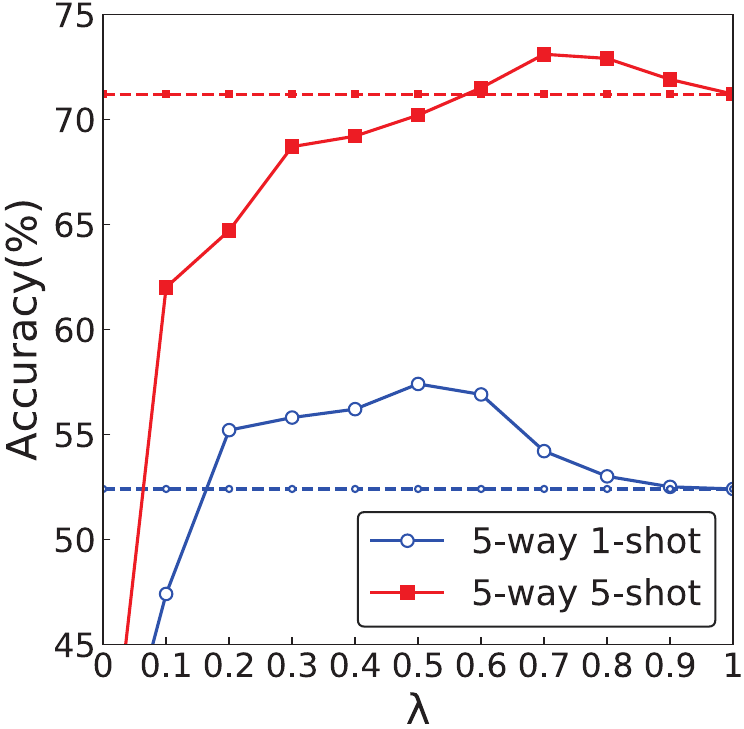}
			\label{fig:2c}
		\end{minipage}
	}
	\caption{(a) Images in each row coming from the same domain. (b) Visualizations of the learned representations using t-SNE for 10-way 5-shot tasks on Easy-Office-Home, where all samples are from the same class and shown in different colors to distinguish their domains with the black point representing the corresponding task-shared prototype. (c) The sensitivity w.r.t. $\lambda$.}
	\label{fig:2}
\end{figure*}

\textbf{The Influence of Data Shift:} In order to further analyze the influence of data shift, we focus on the experimental results on Easy-Office-Home and Hard-Office-Home. 
\begin{itemize}
    \item[(1)] The results of ProtoNet and RelationNet under src+tgt setting are better than those under tgt setting, which reflects that sometimes using more data could achieve better results even when data shift exists. 
    \item[(2)] Several interesting observations jointly show that few-shot learning approaches designed without consideration of data shit have  performance downgrade when data distribution varies.  First, MatchNet achieves relatively worse performance under src+tgt setting than under tgt setting. Second, RelationNet produces better results than ProtoNet on most of the popular benchmarks. In contrast, ProtoNet significantly outperforms RelationNet on our constructed datasets where serious data shift problem exists. Third, the margin between MatchNet~\cite{snell2017prototypical} and ProtoNet~\cite{snell2017prototypical} performed on our constructed datasets is larger than that performed on benchmarks. 
    \item[(3)] Note that the training set in fulltgt setting is smaller than that in src+tgt setting. However, for all the models on all tasks, the performance under fulltgt setting is better than that under src+tgt setting. The main reason is that there is almost no data shift under fulltgt setting but with significant data shift under src+tgt setting. In other words, the existence of data shift does have a negative impact on the performance of few-shot learning models. 
    \item[(4)] The results of transfer methods ProtoNet + DAN, + DANN, and + D-CORAL are worse than the one of the basic methods ProtoNet~\cite{snell2017prototypical}, which indicates that the simple combination of ProtoNet and transfer learning algorithms cannot work well for few-shot under data shift. Meanwhile, the proposed GPN under src+tgt setting not only outperforms ProtoNet in the same setting, but also achieves comparable performance with ProtoNet under fulltgt setting. This phenomenon demonstrates that the data from different distributions can be utilized more effectively by solving the few-shot under data shift problem.
    \item[(5)] The improvement of our GPN on 1-shot is larger than 5-shot. This is mainly because 5-shot learning has more support points, and the task-specific information is more reliable than 1-shot learning. In other words, the influence of data shift is probably smaller in 5-shot than 1-shot setting.
    \item[(6)] On most tasks, our GPN outperforms KTN, which demonstrates our GPN is more effective for few-shot learning under data shift. In addition, comparing our GPN with KTN, it is easy to find that the improvement in 10-way tasks is larger than 5-way tasks. This is mainly because KTN does not follow the meta-learning framework. With the standard supervised framework, KTN learns to classify the training classes with ignoring the test classes, so the generalization ability of the model is not good enough. Thus, with more-way tasks (10-way), the disadvantage of KTN is more obvious, this leads to the worse performance of 10-way tasks in Table~\ref{tab:easyOfficeHome} and~\ref{tab:hardOfficeHome} than 5-way tasks. On the contrary, our GPN utilizes the meta-learning framework. In the training step, each task (iteration) is a few-shot problem, and the model learns to classify any N categories (N-way K-shot problem), which also means that the model learns the ability to classify any N categories. Thus, comparing with KTN, our GPN could have a larger improvement in 10-way tasks than 5-way tasks. This is also the advantage of the meta-learning framework and the reason for many few-shot approaches taking the meta-learning framework. Note that the framework of KTN is more suitable for full-predict few-shot tasks, and most full-predict few-shot methods~\cite{hariharan2017low,gidaris2018dynamic} exploit a similar framework as KTN.
\end{itemize}

\subsection{Analysis}
\textbf{Ablation Study:}
To further analyze the effectiveness of joint training of GCN and the semantic embedding input of GCN, we conduct ablation study on the Easy-Office-Home dataset by evaluating several models (shown in Tabel~\ref{ablation}): (1) GPN (rand): using random vectors as the input of GCN; (2) GPN (fc): using a two-layer  fully connected network, which has 2048 latent nodes and 512 output nodes (i.e., the same as GCN), to replace GCN. As shown in Table~\ref{ablation}, GPN (rand) has a lower average accuracy than GPN, which indicates the importance of introducing semantic embedding information. Besides, GPN achieves better performance than GPN (fc), which demonstrates that GCN can transfer more knowledge from training categories to test categories with the help of knowledge graph.

\textbf{Feature Visualization:}
In Figure~\ref{fig:2b}, we visualize the learned representations of samples from the same class in the 10-way 5-shot task on Easy-Office-Home using t-SNE~\cite{donahue2014decaf}. Different colors represent different domains and the black point denotes the corresponding task-shared prototype learned from GCN. In Figure~\ref{fig:2b}, it can be observed that the distributions of samples with the same label but from different domains are quite different. The existence of data shift in the experiments is also evidenced by this figure. In addition, we can find that the task-shared prototype learned by GCN is about in the center of all points, which further confirms that the proposed GPN is able to learn the task-specific and the task-shared representations.

\textbf{Parameter Sensitivity:}
The parameter $\lambda$ balances the task-shared and the task-specific information. 
Its sensitivity is evaluated on 5-way 1-shot tasks and 5-way 5-shot tasks on Easy-Office-Home by setting different values of $\lambda$, i.e., $\lambda \in \{0, 0.1, \cdots, 0.9, 1\}$.  
Based on Figure~\ref{fig:2c}, we can have the following findings. (\textbf{1}) In the case that only the task-shared representations are used (i.e., $\lambda = 0$), the corresponding performance is very poor, which indicates the necessity of learning task-specific representations. Note that the situation where $\lambda = 0$ is also known as zero-shot problem, and the resultant performance is in line with expectations.  (\textbf{2}) The best performance is not achieved when $\lambda = 1$, which reflects that the task-shared information could help the task-specific representations to become more stable. (\textbf{3}) The 5-way 1-shot and 5-way 5-shot tasks achieve the best performance when $\lambda$ is set to about 0.5 and 0.7, respectively. The reason is that the 5-way 1-shot task has a single support point which is likely to be disturbed. Therefore, additional information from the task-shared representations is supposed to be utilized. In this case, the $\lambda$ should be set to a smaller value. In the 5-way 5-shot task, there are more support points, and the task-specific information is more reliable. Therefore, the value of $\lambda$ should be larger.
\section{Conclusion}
For few-shot learning, most existing approaches focus on how to exploit the limited labeled samples effectively but ignoring the data shift problem. In practice, the data shift problem widely exists in the popular few-shot benchmarks and negatively impacts the existing approaches. In this paper, we have analyzed the data shift problem presented as intra-task and inter-task data shift. In order to solve this problem, we proposed a model called GPN, which learns and considers both the task-shared and the task-specific representations and then produces a more stable representation. Finally, extensive experiments have demonstrated the weakness of some traditional few-shot approaches as well as the effectiveness of the proposed GPN in the data shift scenario.

\section{Acknowledgments}
\label{sec:ack}
The research work is supported by the National Natural Science Foundation of China under Grant No. 62176014, U1836206, 61773361, U1811461.

\bibliographystyle{fcs}
\bibliography{ref}

\begin{biography}{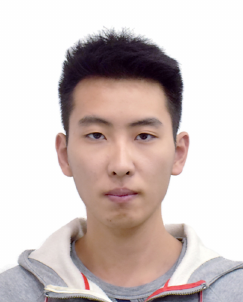}
{Yongchun Zhu}
is currently pursuing his M.S. degree in the Institute of Computing Technology, Chinese Academy of Sciences, Beijing, China. He has published more than 10 papers in journals and conference proceedings including KDD, AAAI, WWW, SIGIR and so on. He received his B.S. degree from Beijing Normal University, China in 2018. His main research interests include transfer learning, meta learning and recommendation system.
\end{biography}

\begin{biography}{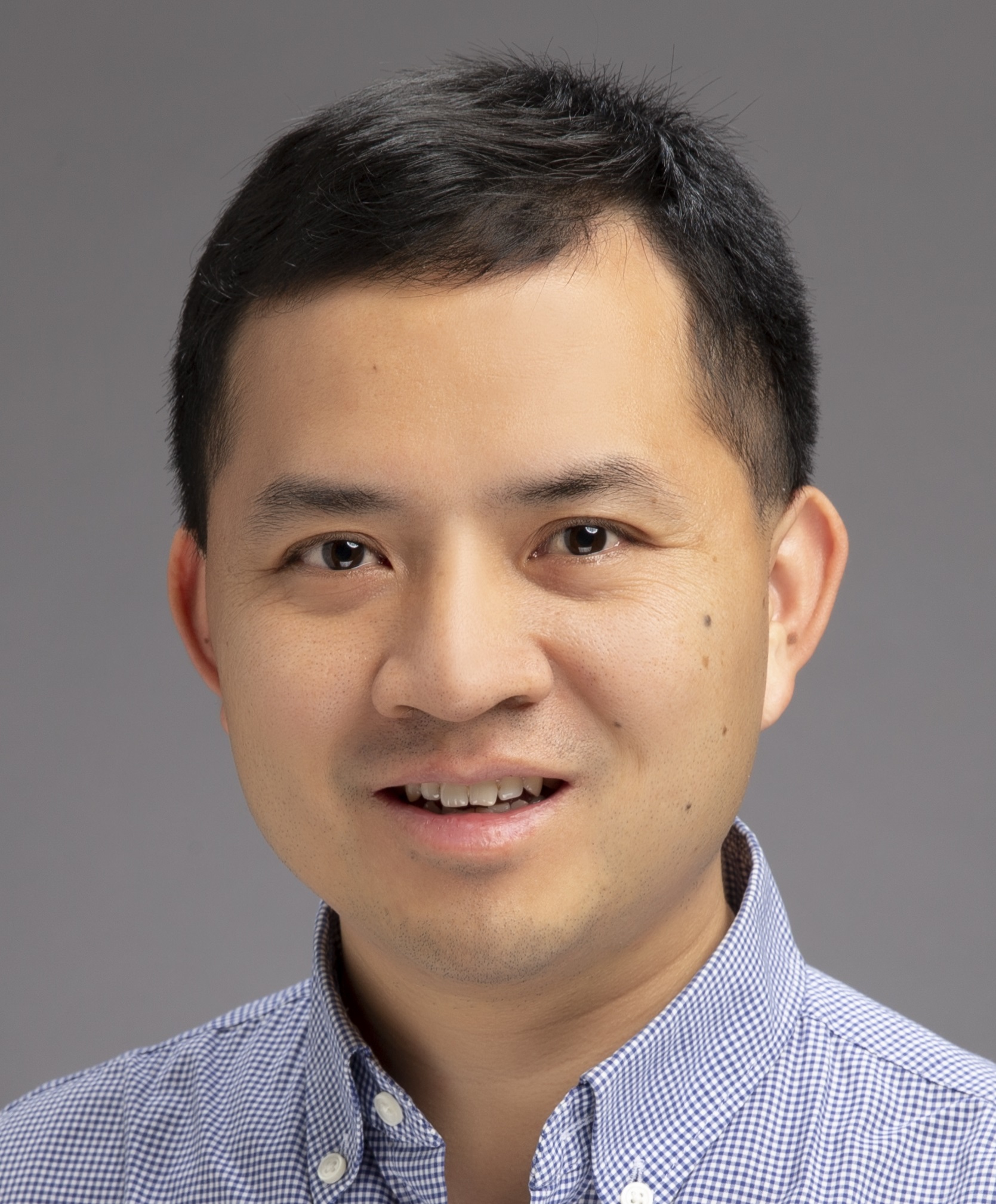}
{Fuzhen Zhuang}
is a professor in Institute of Artificial Intelligence, Beihang University. His research interests include transfer learning, machine learning, data mining, multi-task learning and recommendation systems. He has published more than 100 papers in the prestigious refereed journals and conference proceedings, such as IEEE TKDE, IEEE Transactions on Cybernetics, IEEE TNNLS, ACM TIST, SIGKDD, IJCAI, AAAI, WWW, and ICDE. 
\end{biography}

\begin{biography}{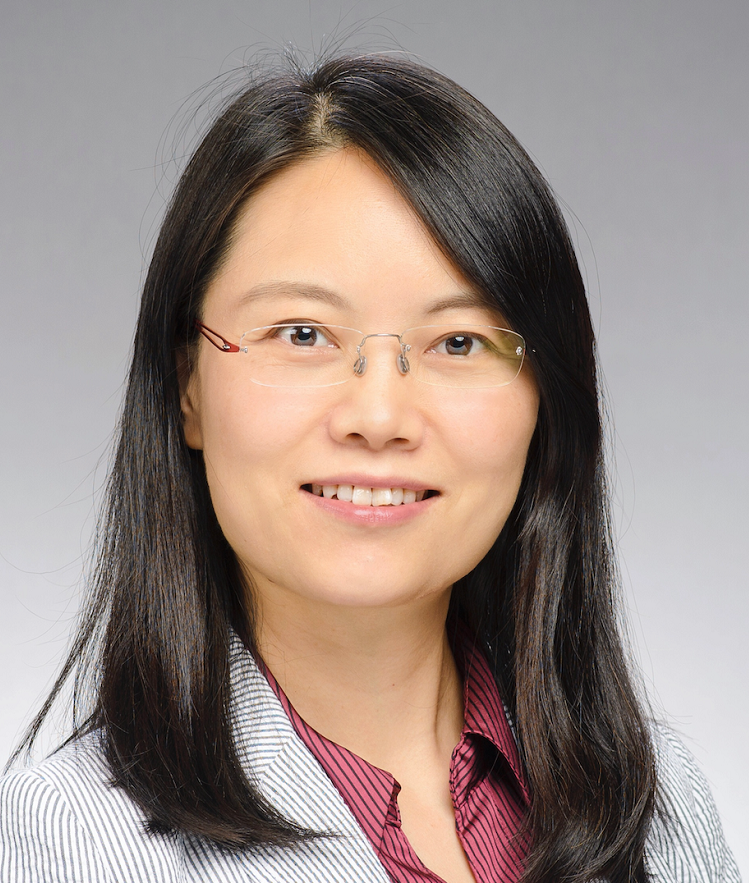}
Xiangliang Zhang is currently an Associate Professor and directs the Machine Intelligence and Knowledge Engineering (MINE) Laboratory at the Department of Computer Science and Engineering in University of Notre Dame, USA.  She received the Ph.D. degree in computer science from INRIA-University  Paris-Sud, France, in July 2010. She has authored or co-authored over 170 refereed papers in various journals and conferences. Her current research interests lie in designing machine learning algorithms for learning from complex and large-scale streaming data and graph data.
\end{biography}

\begin{biography}{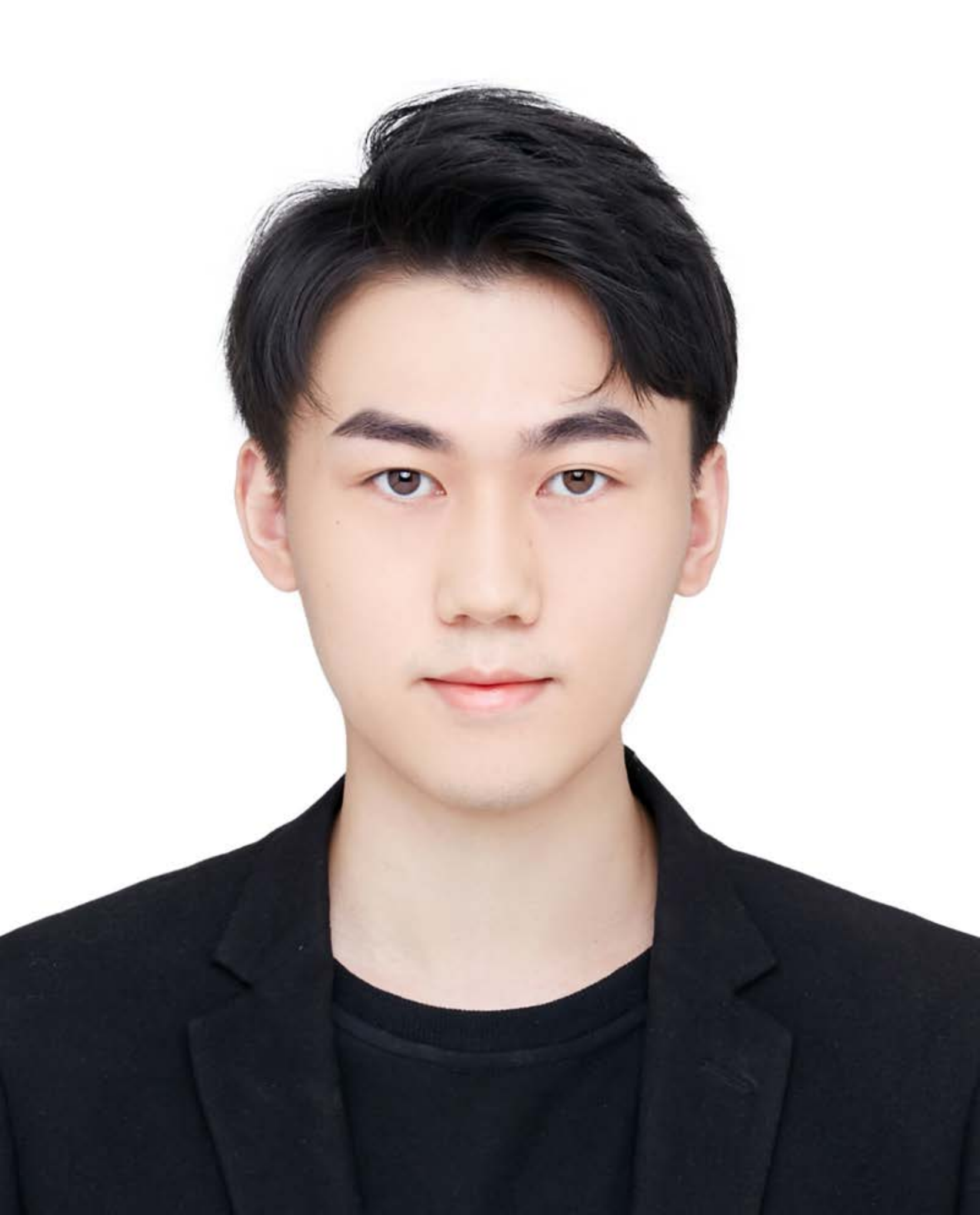}
{Zhiyuan Qi} is currently pursuing his M.S. degree in the University of California, San Diego. He received the B.E. degree in software engineering from Sun Yat-sen University, Guangzhou, China in 2019. He has published several papers in journals and conference proceedings including Proceedings of the IEEE, IEEE Computational Intelligence Magazine, Neurocomputing, IJCAI, etc.
\end{biography}

\begin{biography}{pic/FCS-Shizhiping-fig10.jpg}
{Zhiping Shi} is currently a professor in the College of Information Engineering at the Capital Normal University, Beijing, China. From 2005 to 2010, he was on the faculty at the Institute of Computing Technology, Chinese Academy of Sciences where he received his Ph.D. degree in Computer Software and Theory in 2005. His research interests include formal verification and visual information analysis. He is the (co-)author of more than 100 research papers. He is a Member of the IEEE and the ACM.
\end{biography}

\begin{biography}{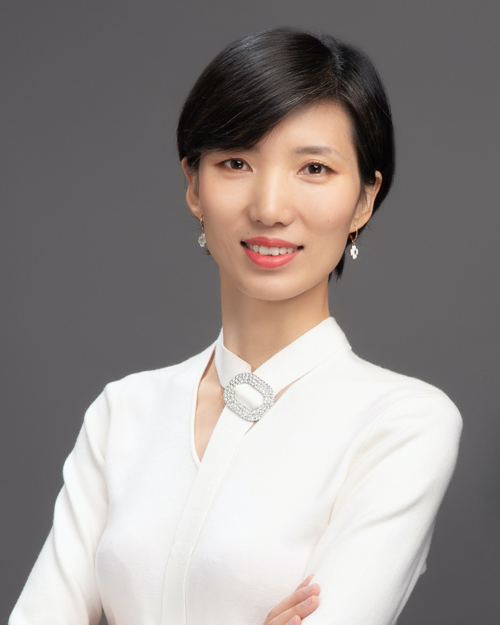}
{Juan Cao} received the Ph.D. degree from the Institute of Computing Technology, Chinese Academy of Sciences, Beijing, China, in 2008. She is currently working as an Professor with the Institute of Computing Technology, Chinese Academy of Sciences. Her research interests include multimedia content analysis, fake news detection, and forgery detection.
\end{biography}

\begin{biography}{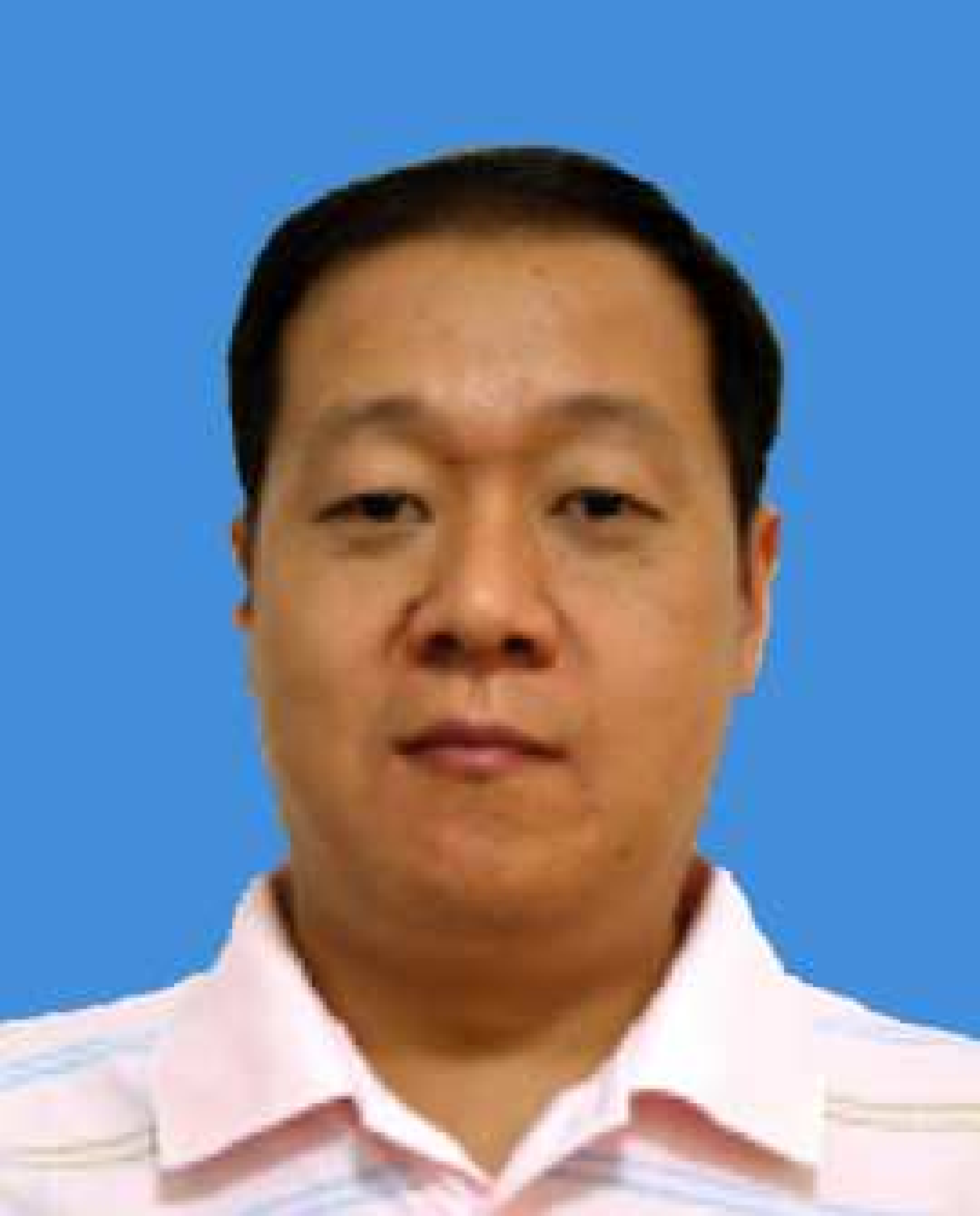}
{Qing He}
is a Professor in the Institute of Computing Technology, Chinese Academy of Science (CAS), and he is a Professor at the Graduate University of Chinese (GUCAS). He received the B.S degree from Hebei Normal University, Shijiazhang, P. R. C., in 1985, and the M.S. degree from Zhengzhou University, Zhengzhou, P. R. C., in 1987, both in mathematics. He received the Ph.D. degree in 2000 from Beijing Normal University in fuzzy mathematics and artificial intelligence, Beijing, P. R. C. Since 1987 to 1997, he has been with Hebei University of Science and Technology. He is currently a doctoral tutor at the Institute of Computing and Technology, CAS. His interests include data mining, machine learning, classification, fuzzy clustering. 
\end{biography}

\end{document}